\documentclass[letterpaper]{article} 
\usepackage[preprint]{aaai2027}  
\usepackage[hyphens]{url}  
\usepackage{graphicx} 
\usepackage{natbib}  
\usepackage{caption} 
\usepackage{algorithm}
\usepackage{algorithmic}

\usepackage{newfloat}
\usepackage{listings}
\DeclareCaptionStyle{ruled}{labelfont=normalfont,labelsep=colon,strut=off} 
\floatstyle{ruled}
\newfloat{listing}{tb}{lst}{}
\floatname{listing}{Listing}

\usepackage{booktabs}
\usepackage{url}            
\usepackage{booktabs}       
\usepackage{amsfonts}       
\usepackage{nicefrac}       
\usepackage{microtype}      
\usepackage{xcolor}         
\usepackage{enumitem}
\usepackage{bbding}
\usepackage{array}
\newcolumntype{L}[1]{>{\raggedright\let\newline\\\arraybackslash\hspace{0pt}}m{#1}}
\newcolumntype{C}[1]{>{\centering\let\newline  \\\arraybackslash\hspace{0pt}}m{#1}}
\newcolumntype{R}[1]{>{\raggedleft\let\newline \\\arraybackslash\hspace{0pt}}m{#1}}
\usepackage{graphicx}
\usepackage{subcaption}
\usepackage{url}
\usepackage{bm}
\usepackage{amsmath,amssymb,amsfonts,amsthm}
\usepackage{caption} 
\usepackage{float} 
\usepackage{verbatim}
\usepackage{algorithm}
\usepackage{algorithmic}
\usepackage{multirow}
\usepackage[most]{tcolorbox}
\usepackage{makecell}

\theoremstyle{definition}
\newtheorem{definition}{Definition}
\title{Falsifiable Commitment Planning for Self-Correcting Web Agents}
\author{
    Guangyi Liu\textsuperscript{\rm 1},
    Huan Zhao\textsuperscript{\rm 2},
    Quanming Yao\textsuperscript{\rm 1}\corresponding
}
\affiliations{
    \textsuperscript{\rm 1}Tsinghua University\qquad
    \textsuperscript{\rm 2}Noumena AI\\
    liugy24@mails.tsinghua.edu.cn, zhaohuan@noumena.com.cn, qyaoaa@tsinghua.edu.cn
}

\newcommand{\method}{\textsc{FCPAgent}}
\newcommand{\fcu}{\textsc{FCU}}
\newcommand{\fcus}{\textsc{FCU}s}
\newcommand{\pos}{E^{+}}
\newcommand{\fals}{E^{-}}
\newcommand{\cmark}{\ensuremath{\checkmark}}
\newcommand{\xmark}{\ensuremath{\times}}

\begin{document}

\maketitle

\begin{abstract}
	Long-horizon web agents often go off track before final failure: a trajectory can remain locally plausible even after the current state, reused skill, or plan assumption no longer supports the user instruction. Existing agents can plan, reflect, or reuse experience, but their plans rarely specify the evidence under which an active step should still be trusted. We propose \method{}, a falsifiable commitment planning framework for robust long-horizon web agents. \method{} represents each plan step as a Falsifiable Commitment Unit (\fcu{}): a subgoal grounded in a reusable skill, together with confirming evidence, falsifying evidence, and a confidence score. Execution is organized as a plan-test-repair loop. The hybrid commitment testing module checks candidate actions before they modify the browser and checks observations after execution; for efficiency, it combines lightweight evidence matching with LLM-based diagnostic verification. When evidence falsifies a commitment, scope-aware repair localizes the contradiction to the execution, skill, or planning level and revises the smallest adequate part. On WebArena, \method{} achieves a 13.8\% relative improvement in average success over the strongest baseline, with especially large gains on long-horizon tasks.
\end{abstract}

\section{Introduction}


Web browsers have become a universal interface for digital work, supporting tasks such as shopping, information search, and enterprise services. Recent LLM- and VLM-based agents can follow natural-language instructions to operate web interfaces through clicks, typing, scrolling, form filling, and tool calls \cite{yao2023react,he2024webvoyager,zheng2024seeact,yang2025agentoccam}, shifting web automation from hand-coded scripts to generalist multi-step agents. Yet reliable deployment remains difficult because web tasks are partially observed, dynamically updated, and often require long chains of dependent decisions.

A key obstacle is off-track execution before final failure. For example, an agent may enter a wrong branch of the website and repeatedly search pages that do not contain the information required. Each subsequent click can still look locally reasonable because the page labels are semantically related to the task, although the trajectory no longer supports the user's instruction. Such trajectory deviation can arise when the next action targets the wrong element, when a retrieved skill no longer fits the current layout or permission state, or when an earlier assumption invalidates the remaining plan. Recent methods improve robustness through progress reflection, dynamic replanning, and skill induction \cite{zhou2026colorbrowseragent,yang2025webdart,li2026skilltracer}, but their plans are usually not formulated as testable objects with explicit validity conditions. Without an explicit way to notice execution drift, an agent may continue following a plausible but stale routine after the active plan has become invalid.

We argue that stable long-horizon web agents need plans that know when they are wrong. Cognitive theories of goal-directed behavior suggest that robust action is inherently anticipatory: agents act with expectations about what should happen next and revise behavior when observations violate those expectations~\cite{miller1960plans}. Instead of treating a plan step as an executable recipe, the agent should treat it as a \emph{falsifiable commitment}: a hypothesis that the plan remains valid and that execution is still progressing along a task-supporting path. Each commitment should indicate evidence for normal progress, as well as evidence that execution may have drifted. In this view, execution becomes online commitment testing: as the browser state evolves, the agent checks whether the active step remains supported or has been contradicted, allowing it to detect drift before final failure and trigger timely correction.

We introduce \method{}, a Falsifiable Commitment Planning framework for long-horizon web tasks. \method{} represents a plan as a sequence of \emph{Falsifiable Commitment Units} (\fcus{}), each binding a subgoal to a reusable skill, expected progress evidence, hierarchical falsifiers, and a confidence score. The framework executes these commitments through a plan-test-repair loop: it generates falsifiable commitments, tests them before and after browser actions, and repairs the smallest violated part when evidence fails. A hybrid tester uses lightweight evidence matching for common confirmations and contradictions, while invoking LLM-based diagnostic verification only for completion-like, risky, or ambiguous states. When a commitment is falsified, an experience-enhanced scope-aware repair module chooses an adequate revision, ranging from local action correction to skill switching, replanning the remaining commitments, or restarting from a known safe state. After execution, \method{} distills successful trajectories into reusable skills and repaired or failed trajectories into failure-repair memories, making future commitments easier to write, test, and repair.
Our contributions are summarized as follows:
\begin{itemize}[leftmargin=*]
    \item We formulate \emph{falsifiable commitment planning}, which
represents each plan step as a testable commitment with explicit
confirming and falsifying evidence, providing a runtime validity
interface between high-level planning and low-level execution.

\item We instantiate this formulation in \method{}, a
plan--test--repair framework that tests commitments before and after
browser actions and uses diagnosed violations to revise the smallest
implicated execution, skill, or planning scope.

\item We demonstrate consistent improvements on WebArena, including
a 13.8\% relative gain over the strongest baseline, and a 9.6\%
relative zero-shot gain on WebChoreArena, with the largest
improvements on long-horizon tasks.
\end{itemize}

\section{Related Work}

\subsection*{LLM-based web agents}
LLM-based web agents have evolved from reactive next-action prediction over textual or DOM observations into integrated systems that combine perception, planning, learning, memory, and tool use. Recent progress has followed several complementary paths: improved observation and action abstractions for browser grounding \cite{he2024webvoyager,zheng2024seeact,yang2025agentoccam}; training-based agents using demonstrations, synthetic trajectories, or reinforcement learning \cite{lai2024autowebglm,qi2025webrl}; world-model-based agents that predict web-state transitions before acting \cite{chae2025wma}; planning, decomposition, and reflection methods that revise task structure during execution \cite{shinn2023reflexion,wang2026preflect,yang2025webdart}; and memory or skill-augmented agents that reuse prior interaction knowledge \cite{wang2025awm,prabhu2026walt,zhou2026colorbrowseragent}. Across these directions, long-horizon reliability remains difficult because local action validity does not guarantee that the active plan remains valid. 

\subsection*{Planning and reflection}
Planning and reflection have been widely studied in LLM agents, such as Tree of Thoughts, Reflexion, and Self-Refine \cite{yao2023tot,shinn2023reflexion,madaan2023selfrefine}. More recent web-agent methods make planning more adaptive: 
Anticipatory Reflection (AR) anticipates potential action failures before execution by proposing backup actions that can be tried upon subsequent misalignment \cite{wang2024devils};
WebDART decomposes web tasks into navigation, extraction, and execution subtasks and updates the decomposition online \cite{yang2025webdart}; ColorBrowserAgent uses progress summarization and environment adaptation to reduce trajectory deviation in long-horizon browsing \cite{zhou2026colorbrowseragent}. 
These studies show that long-horizon web execution benefits from explicit task structure, progress assessment, and adaptive revision during interaction.

\subsection*{Skills and memory for agents}
Skill and memory mechanisms reduce repeated reasoning by reusing prior experience, as in Voyager~\cite{wang2023voyager}, Agent Workflow Memory~\cite{wang2025awm}, SkillWeaver~\cite{zheng2025skillweaver}, ASI~\cite{wang2025asi}, Hierarchical Memory Tree~\cite{tan2026hmt}, and ContractSkill~\cite{lu2026contractskill}. 
WALT~\cite{prabhu2026walt} learns deterministic tools from websites, turning recurring site functions such as search or filtering into callable procedures. 
SkillTracer~\cite{li2026skilltracer} represents composite skills as attributed plan graphs and localizes long-horizon failures to specific structural components, enabling targeted refinement rather than discarding the whole skill.

\begin{figure}[t]
    \centering
    \includegraphics[width=1.0\columnwidth]{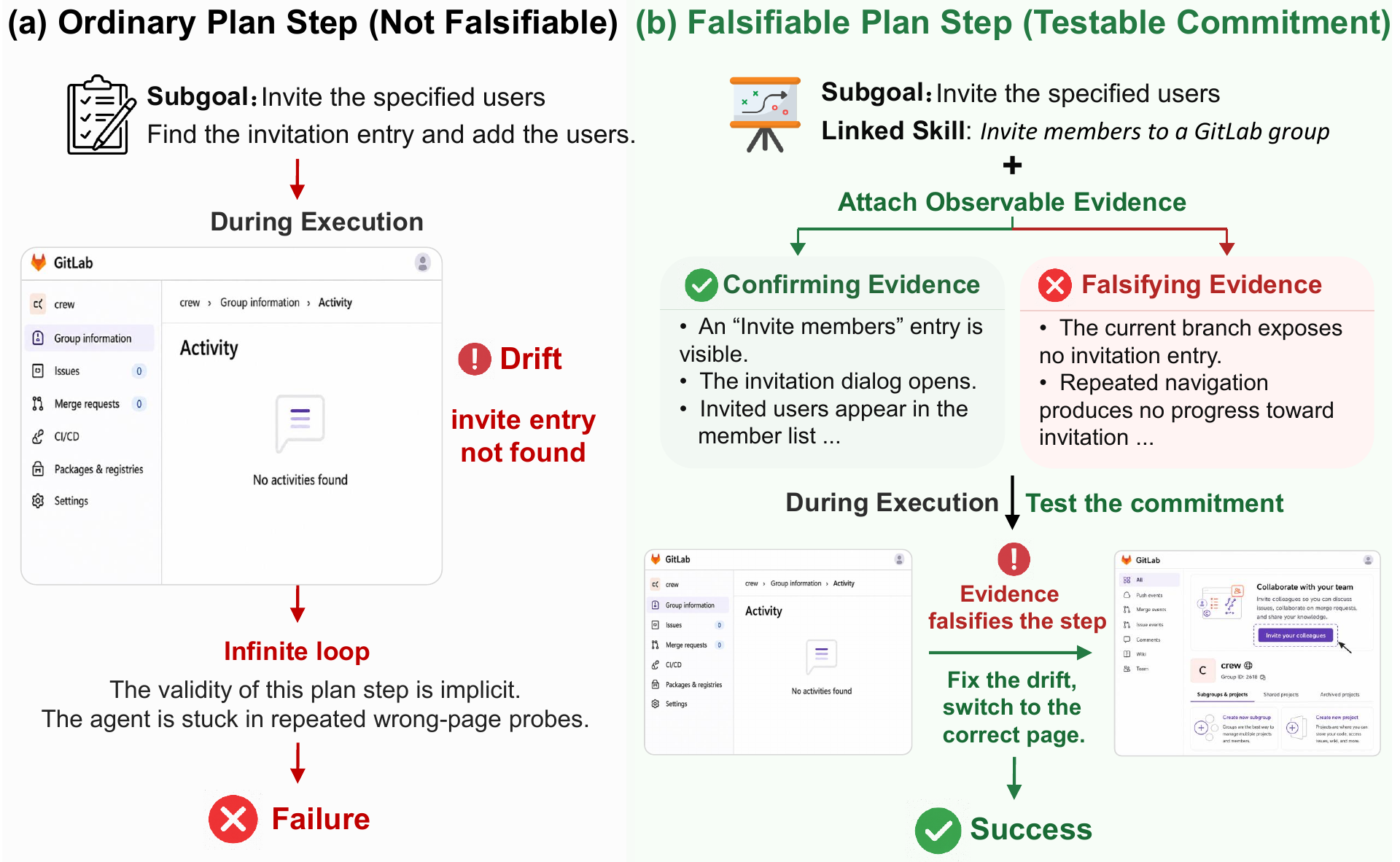}
    \caption{An off-track execution example. An ordinary plan leaves its runtime validity implicit and may loop on wrong pages. A falsifiable plan step attaches confirming and falsifying evidence, allowing \method{} to detect the drift, repair the navigation path, and complete the task successfully.}
    \label{fig:fcu}
\end{figure}

\begin{figure*}[ht]
	\centering
	\includegraphics[width=.96\textwidth]{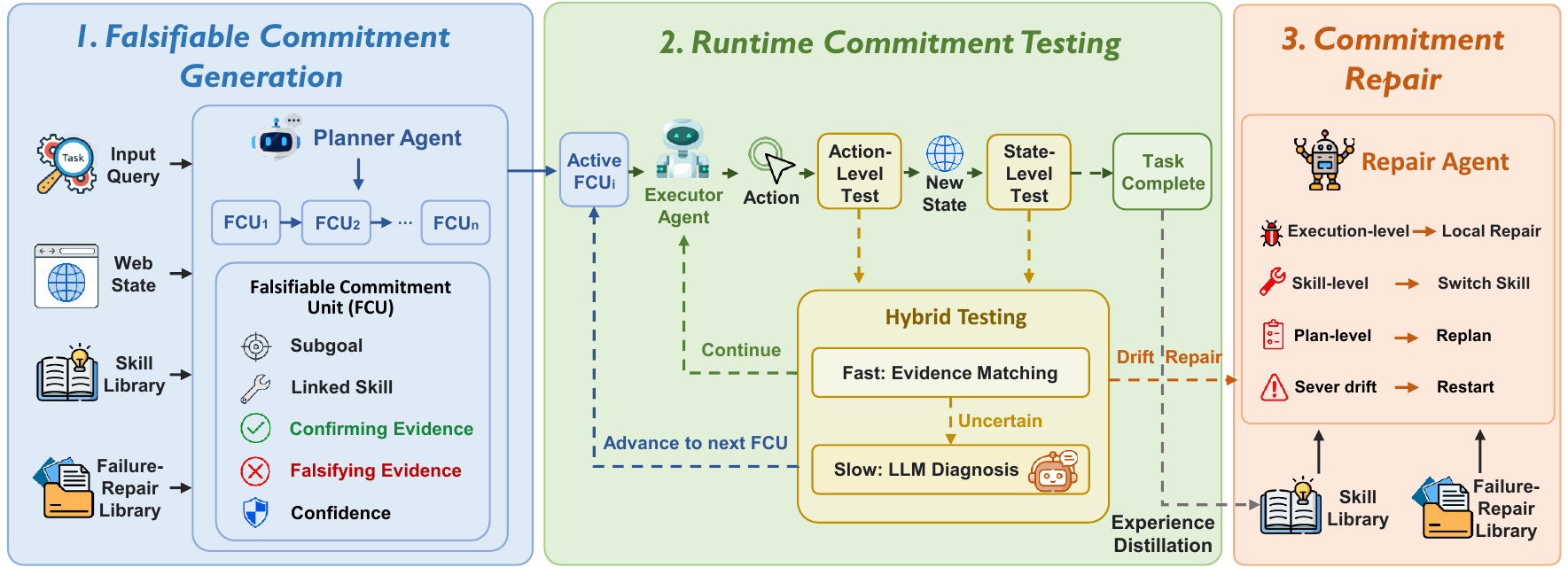}
	\caption{\method{} casts web-agent execution as online testing of falsifiable commitments. Hybrid commitment testing checks actions and observations; scope-aware repair revises the smallest contradicted component.}
	\label{fig:method}
\end{figure*}

\section{Motivational Example}

Figure~\ref{fig:fcu} illustrates a GitLab task that requires creating a group and inviting several specified users. After creating the group, an agent may enter the \emph{Group information} branch while searching for the invitation entry. The invitation subgoal itself remains valid, and entering this branch appears locally plausible given its semantic relevance to the task. However, because the invitation control is available only from the group overview page, the agent may spend many steps exploring nearby pages without making progress. We call this failure mode \emph{off-track execution}: locally reasonable actions continue even though the active state or plan step no longer supports task completion.

Existing web agents can plan, reflect, and reuse previously learned skills, but the runtime validity of an active plan step is usually left implicit. A step such as \emph{invite members} states the intended operation, but does not specify what observable evidence indicates progress or what observation should make the agent stop trusting the current path. Because final task feedback is delayed and sparse, the agent lacks an operational boundary between acceptable execution variation and meaningful drift.

Inspired by the Test--Operate--Test--Exit view of goal-directed behavior~\cite{miller1960plans}, we treat each plan step as a provisional commitment whose validity should be tested as the execution situation evolves. A falsifiable plan step associates its intended operation with observable evidence that either confirms or falsifies its continued validity. In the GitLab example, entering group-related pages without exposing an invitation entry satisfies an execution-level falsifier. The agent can then return to the group overview, find the correct invitation entry, and complete the task. By making such violations explicit, falsifiable commitment planning enables agents to correct themselves before local drift compounds, improving stability over long-horizon execution.

\section{Problem Formulation}

A web task is specified by a natural-language instruction $q$. At step $t$, the agent observes a browser state $o_t$, which may include an accessibility tree and screenshots, selects an action $a_t\in\mathcal{A}$, and receives a new observation $o_{t+1}\sim\mathcal{T}(o_t,a_t)$. The resulting trajectory is $\tau_T=(o_0,a_0,o_1,\ldots,a_{T-1},o_T)$, and task success is evaluated by a validator $V(q,\tau_T)\in\{0,1\}$. Since this final feedback is delayed and sparse, the agent requires intermediate evidence for determining whether its active plan remains valid.

\begin{definition}[Falsifiable Commitment Unit]
	\label{def:fcu}
	A Falsifiable Commitment Unit (\fcu{}) for the $i$-th plan step is
	\begin{equation}
		c_i=(g_i,s_i,\pos_i,\fals_i,\kappa_i),
		\label{eq:fcu}
	\end{equation}
	where $g_i$ is the current subgoal, $s_i$ is an optionally linked reusable 	skill, $\pos_i$ and $\fals_i$ specify confirming and falsifying evidence, and $\kappa_i\in[0,1]$ denotes commitment confidence. The unit represents a provisional and testable assertion that pursuing $g_i$, optionally
	through $s_i$, remains appropriate at the current execution stage.
\end{definition}

Unlike an ordinary plan step that only specifies what to do, an \fcu{} also specifies what should be observed if the step remains valid and what evidence should make the agent stop trusting it. Operationally, falsifiable commitment planning follows the principle
\begin{equation}
c_i
\!\xrightarrow{\text{runtime evidence}}\!
\begin{cases}
	\text{proceed}, & \!\!\text{if } c_i \text{ remains supported},\\
	\text{repair},  & \!\!\text{if } c_i \text{ is falsified}.
\end{cases}
\label{eq:fcp-principle}
\end{equation}
Here, \emph{proceed} means either continuing the active plan step or advancing after its completion is confirmed, whereas \emph{repair} means revising the implicated part of the current execution. Thus, a commitment is not followed unconditionally; it is trusted only while runtime evidence continues to support it.

The confirming evidence $\pos_i$ is organized stage-wise: precondition evidence checks whether the current page supports the subgoal, progress evidence checks whether execution follows an expected path, and completion evidence checks whether the subgoal has been achieved. The falsifying evidence $\fals_i$ is organized hierarchically: execution-level falsifiers indicate local action or state drift, skill-level falsifiers indicate that the selected skill no longer fits the current site state, and planning-level falsifiers indicate that the subgoal or decomposition itself is suspect. 
Confirming evidence allows the agent to proceed while the commitment remains supported, whereas falsifying evidence indicates both when correction is needed and whether the execution, skill, or plan should be reconsidered. Together, they provide a shared diagnostic interface for planning, testing, and repair.

\section{The Proposed Method}

The formulation above specifies what a monitorable plan step should contain, but leaves three operational questions open:
\begin{itemize}
\item \emph{how to construct task-specific commitments};
\item \emph{how to test their validity during execution}; and
\item \emph{how to revise when a commitment is falsified}.
\end{itemize}
As illustrated in Figure~\ref{fig:method}, the following three subsections answer these questions in the same order. 
\emph{Falsifiable Commitment Generation} uses reusable skills and past failure-repair experience to construct a compact commitment sequence $C=(c_1,\ldots,c_m)$. 
\emph{Runtime Commitment Testing} examines both proposed actions and resulting browser states to determine whether the active commitment should continue, advance, or require repair. \emph{Repairing Falsified Commitments} then localizes a detected violation and revises the execution, skill, or planning scope as needed. Together, these components form the plan--test--repair loop of \method{}.

\subsection{Falsifiable Commitment Generation}
The first question is how to construct task-specific commitments that are both procedurally grounded and risk-aware. Commitments generated only from the task instruction may contain evidence that is too generic to guide browser execution, whereas commitments copied directly from reusable skills may overlook task-specific constraints and known failure modes. \method{} addresses this question by using successful experience as procedural priors and failure-repair experience as guardrails.

Before test-time execution, \method{} distills successful training trajectories into a skill library $\mathcal{S}$ and failed or repaired trajectories into a failure-repair library $\mathcal{F}$. Skills store reusable procedures and success conditions, whereas failure memories store failure patterns and repair lessons. Both libraries are built only from training trajectories and remain fixed during test-time evaluation.

Given the task query $q$ and initial browser state $o_0$, the planner retrieves relevant skills $\mathcal{S}_q\subseteq\mathcal{S}$ and failure-repair memories $\mathcal{F}_q\subseteq\mathcal{F}$, then generates a compact commitment sequence:
\begin{equation*}
C=\operatorname{Planner}\big(q,o_0,\mathcal{S}_q,\mathcal{F}_q\big).
\end{equation*}
Each commitment $c_i$ specifies a subgoal $g_i$, an optional linked skill $s_i$, confirming evidence $\pos_i$, falsifying evidence $\fals_i$, and confidence $\kappa_i$. 
The retrieved skills provide candidate procedures and observable success signals for recurring subgoals. The planner may link an \fcu{} to a suitable skill, grounding its subgoal and confirming evidence in successful experience. Failure-repair memories expose known risks before execution starts, allowing the planner to encode anticipated failure patterns as falsifiers. Thus, skills help specify what should work, while failure memories help specify what would make a commitment suspect.


\subsection{Runtime Commitment Testing}

The second question is how to determine whether an active commitment remains supported during execution. \method{} answers this question through hybrid commitment testing: a lightweight tester frequently screens candidate actions and returned browser states, while an LLM-based verifier performs deeper diagnosis only when needed. This design detects local drift without requiring an expensive LLM judgment after every browser transition.

\subsubsection{Lightweight Action- and State-Level Testing}

At time $t$, the executor proposes an action for the active commitment $c_i$. Action-level testing checks the proposal before it modifies the browser, while state-level testing checks whether the returned observation $o_{t+1}$ supports or contradicts $c_i$.

For state-level testing, let $T_t$ and $I_t$ denote the accessibility tree and screenshot, and let $E_i^{\pm}$ denote either confirming evidence $\pos_i$ or falsifying evidence $\fals_i$. After rule-based anomaly checks, the lightweight tester computes both scores with
\begin{equation*}
\alpha_t^{\pm}=\lambda\,\mathrm{NLI}(T_t,E_i^{\pm})+(1-\lambda)\,\mathrm{ITM}(I_t,E_i^{\pm}),
\end{equation*}
where $\lambda$ balances textual entailment and image--text matching, instantiated with \texttt{nli-deberta-v3-base} and \texttt{SigLIP2}, respectively. In pre-action testing, $T_t$ is replaced by the candidate reasoning--action pair and the visual term is omitted.

The slow tester is invoked when the completion component of $\pos_i$ produces a strong positive score, $\fals_i$ produces a strong falsification score, or the two scores conflict or lie near the routing thresholds; rule-detected execution anomalies are also escalated. Thus, completion-like, risky, or ambiguous cases are escalated to the LLM verifier, whereas routine cases with weak evidence continue directly. Confidence $\kappa_i$ further adjusts the routing sensitivity.

\subsubsection{LLM-Based Diagnostic Verification}

For each escalated case, the LLM verifier examines the task $q$, active commitment $c_i$, recent trajectory $\tau_{\leq t}$, relevant action or observation, and the lightweight tester's routing signal. The lightweight scores serve only as routing cues rather than hard decisions. By reasoning over the recent trajectory and current browser context, the verifier distinguishes acceptable execution variation from genuine falsification and identifies whether the contradiction is execution-level, skill-level, or planning-level. It returns
\[
d_t \in \{\textsc{continue},\textsc{advance},\textsc{repair}\},
\]
together with a diagnosis $u_t$ of the falsifying evidence pattern and its scope when $d_t=\textsc{repair}$. A \textsc{continue} decision proceeds with $c_i$; \textsc{advance} confirms completion, moves to the next commitment, and compresses the previous interactions into a short stage summary to keep the context compact; \textsc{repair} passes $u_t$ to the scope-aware repair module.


\subsection{Repairing Falsified Commitments}
\label{sec:repair}

The third question is how to revise the appropriate part after a commitment is falsified without discarding unaffected progress. When the verifier returns $d_t=\textsc{repair}$, the diagnosis $u_t$ identifies the evidence that contradicts the active commitment $c_i$. This attribution matters because the same symptom, such as a missing control, may arise from an incorrect action, an unsuitable skill, or an invalid plan and therefore require different repairs. \method{} answers this question through experience-enhanced scope-aware repair: relevant past failures suggest possible recovery strategies, while the diagnosed falsifier determines the smallest component that must be revised.

At repair time, the repairer retrieves cases from the failure-repair library $\mathcal{F}$ using both the task $q$ and diagnosis $u_t$, matching task similarity and failure cause, respectively. Retrieved cases provide strategies and lessons from analogous failures but serve only as soft guidance: the repair remains grounded in the active commitment $c_i$, recent trajectory $\tau_{\leq t}$, current browser state, and retrieved skills $\mathcal{S}_q$.

\subsubsection{Scope-Aware Commitment Revision}

Let $C=(c_1,\ldots,c_i,\ldots,c_m)$ be the current sequence. The repairer preserves the completed prefix $(c_1,\ldots,c_{i-1})$. Execution-level falsifiers preserve $c_i$ and revise the immediate action or path; skill-level falsifiers preserve $g_i$ but replace $s_i$ and update the affected evidence; planning-level falsifiers rewrite $c_i$ and the affected suffix. This hierarchy expands the revision scope only when a smaller correction is inadequate.

Revision scope and recovery depth are determined separately. The diagnosed falsifier determines what should change, whereas state severity determines whether execution resumes locally or first backtracks to a known safe state. The revised sequence is then returned to the executor, completing the plan--test--repair loop of \method{}.

Further implementation details, including skill and failure-repair record formats, experience retrieval and induction, and hybrid testing and routing, are provided in the Appendix.


\subsection{Comparison with Related Work}

\begin{table}[t]
	\centering
	\begin{tabular}{lcccc}
        \toprule
        Method
        & ETPC
        & EDD
        & MSFA
        & AGR
        \\
        \midrule
        AgentOccam& \xmark & \xmark & \xmark & \xmark \\
        AWM& \xmark & \xmark & \xmark & \xmark \\
        WebDART& \xmark & \cmark & \xmark & \xmark \\
        WALT& \xmark & \xmark & \xmark & \xmark \\
        ContractSkill& \xmark & \cmark & \xmark & \cmark \\
        ColorBrowserAgent& \xmark & \cmark & \xmark & \xmark \\
        \textbf{FCPAgent (Ours)}& \cmark & \cmark & \cmark & \cmark \\
        \bottomrule
	\end{tabular}
	\caption{Comparison along four long-horizon reliability dimensions. ETPC: explicit task-progress criteria; EDD: explicit drift detection; MSFA: multi-scope failure attribution; AGR: attribution-guided recovery.}
	\label{tab:compare}
\end{table}

Table~\ref{tab:compare} organizes representative methods according to four general requirements for reliable long-horizon execution. Several methods provide explicit runtime monitoring: WebDART updates its task decomposition when execution no longer supports the current plan, while ColorBrowserAgent combines progress assessment with environment adaptation \cite{yang2025webdart,zhou2026colorbrowseragent}. ContractSkill further localizes violations within structured skill representations and perform targeted skill-level repair \cite{lu2026contractskill}. \method{} differs in the scope of the monitored object and diagnosis: it specifies task-specific progress and falsification criteria for each active plan step, attributes violations across execution, skill, and planning scopes, and uses the attributed source to select recovery. Its distinction therefore lies not in drift detection or localized repair alone, but in unifying task-specific validity criteria, multi-scope attribution, and attribution-guided recovery.
Detailed comparisons are provided in the Appendix.

\begin{table*}[ht]
	\centering
	\begin{tabular}{lccccccc}
		\toprule
		\textbf{Method} & Shopping & Admin & GitLab & Reddit & Map & Cross & Avg. \\
		\midrule
		ReAct~\cite{yao2023react} & 30.2 & 29.1 & 22.2 & 24.2 & 26.3 & 18.9 & 26.2 \\
		AWM~\cite{wang2025awm} & 40.3 & 39.0 & 31.9 & 51.9 & 30.3 & 21.6 & 37.3 \\
		AgentOccam~\cite{yang2025agentoccam} & 48.9 & 54.8 & 53.6 & 48.5 & 32.1 & 22.4 & 47.5 \\
		WebDART~\cite{yang2025webdart} & 44.7 &	47.5&56.8&52.9	&31.3	&21.6&	46.1 \\
		WALT~\cite{prabhu2026walt} & 51.1 & 56.2 & 57.0 & 48.5 & 36.3 & 24.3 & 49.7 \\
		ColorBrowserAgent~\cite{zhou2026colorbrowseragent} & 59.7 & 60.3 & 63.4 & 70.8 & 37.5 & 27.0 & 57.4 \\
		\textbf{\method{} (Ours)} & \textbf{67.6} & \textbf{68.8} & \textbf{75.5} & \textbf{75.3} & \textbf{41.3} & \textbf{33.2} & \textbf{65.3} \\
		\midrule
		Relative improvement & 13.2\% & 14.1\% & 19.1\% & 6.4\% & 10.1\% & 23.0\% & 13.8\% \\
		\bottomrule
	\end{tabular}
	\caption{Performance comparison on WebArena. The metric is success rate (\%).}
	\label{tab:main}
\end{table*}

\section{Experiments}

\subsection{Setup}

\textbf{Benchmark.} We evaluate on WebArena, which provides functional success validators over 6 realistic web applications \cite{zhou2024webarena}. The main metric is task success rate. Following prior work~\cite{zhou2026colorbrowseragent}, we use OpenStreetMap as the live map environment for map-related tasks.
Dataset statistics and the evaluation protocol are provided in the Appendix.

\noindent
\textbf{Baselines.} We compare \method{} with ReAct \cite{yao2023react}, AWM \cite{wang2025awm}, AgentOccam \cite{yang2025agentoccam}, WebDART \cite{yang2025webdart}, WALT \cite{prabhu2026walt}, and ColorBrowserAgent \cite{zhou2026colorbrowseragent}. For fairness, the skill and failure-repair libraries are built offline from training trajectories and remain fixed during test evaluation. Baselines that require offline preparation (e.g., AWM~\cite{wang2025awm}) are given the same data budget.

\noindent
\textbf{Implementation details.} 
We use Qwen3.5-397B-A17B as the backbone model \cite{qwen2026qwen35} via API call for all methods, with the temperature set to 1. We typically set the number of retrieved skills $K = 3$, the number of retrieved failure experiences $P = 2$, and the hyperparameter $\lambda = 0.8$.
Additional setup and implementation details are provided in the Appendix.

\subsection{Main Results}

Table~\ref{tab:main} shows that \method{} consistently improves over prior systems. Compared with the strongest baseline, ColorBrowserAgent, \method{} raises average success from 57.4\% to 65.3\%, a 13.8\% relative improvement. The gain is largest on GitLab, where success improves from 63.4\% to 75.5\%, and is also substantial on Shopping and Admin. These results suggest that falsifiable commitments help not only with deep reasoning, but also with keeping reused skills and intermediate page states aligned with the active subgoal. The absolute scores on Map and Cross remain lower than those on other domains. Following prior work, we use OpenStreetMap as a real map website environment; as a result, some measured map attributes can deviate from the benchmark references, making these tasks less stable under automatic validation.

\subsection{Zero-Shot Generalization}

\begin{table}[t]
	\centering
	\setlength{\tabcolsep}{1.5pt}
	\begin{tabular}{lccccc}
		\toprule
		\textbf{Method} & Shopping & Admin & GitLab & Reddit & Avg. \\
		\midrule
		AgentOccam & 17.1 & 25.4 & 21.3 & 12.1 & 19.7 \\
		WebDART & 21.4 & 27.5 & 22.8 & 22.0 & 23.7 \\
		ColorBrowserAgent & 24.8 & 32.6 & 32.2 & 24.2 & 28.9 \\
		\textbf{\method{} (Ours)} & \textbf{28.2} & \textbf{35.5} & \textbf{33.1} & \textbf{28.6} & \textbf{31.7} \\
		\bottomrule
	\end{tabular}
	\caption{Zero-shot generalization performance on WebChoreArena. The metric is success rate (\%).}
	\label{tab:webchorearena}
\end{table}

We further test cross-benchmark generalization on WebChoreArena~\cite{miyai2025webchorearena}, a reproducible WebArena extension with more labor-intensive tasks. We report results on its four site-specific domains and apply \method{} zero-shot, keeping the WebArena-derived skill and repair libraries fixed without using WebChoreArena trajectories.
As shown in Table~\ref{tab:webchorearena}, \method{} ranks first in all domains and achieves a weighted average of 31.7\%, compared with 28.9\% for ColorBrowserAgent. This 9.6\% relative improvement indicates that falsifiable commitment planning and WebArena-derived experience remain effective under a harder task distribution, even when no target-benchmark experience is available.

\begin{figure}[ht]
	\centering
	\includegraphics[width=\columnwidth]{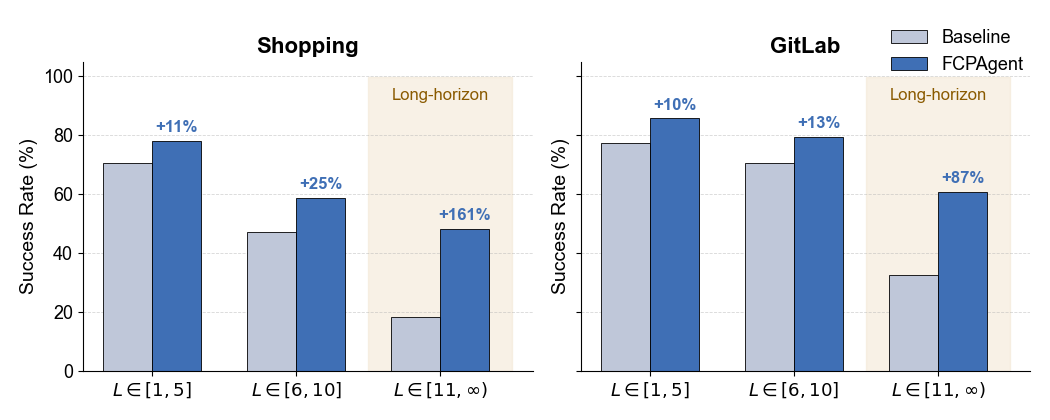}
	\caption{Success rate (\%) by task length. The strongest gains occur in the longest buckets.}
	\label{fig:length}
\end{figure}

\subsection{Long-Horizon Robustness}

The central motivation of \method{} is robustness under long horizons. Figure~\ref{fig:length} groups Shopping and GitLab tasks by the number of interaction steps. Since WebArena provides no oracle interaction length, we use each task's trajectory length under the strongest baseline, ColorBrowserAgent, as a fixed proxy for its execution horizon; every bucket contains more than 30 tasks. \method{} improves over the baseline~\cite{zhou2026colorbrowseragent} in every bucket, but the gains grow sharply with task length. On Shopping, the relative improvement rises from 11\% for 1--5 step tasks to 161\% for tasks with at least 11 steps. On GitLab, the corresponding long-horizon gain is 87\%, much larger than the 13\% gain in the medium-length bucket. These results support the hypothesis that falsifiable commitments are most valuable when small intermediate trajectory deviations would otherwise compound silently over many actions.

\begin{table}[ht]
	\centering
	\begin{tabular}{lccc}
		\toprule
		Variant & Shopping & Admin & GitLab \\
		\midrule
		Full  & \textbf{67.6} & \textbf{68.8} &  \textbf{75.5}\\
		w/o falsifiable plan & 60.4 & 61.7 & 71.2 \\
		w/o commitment testing & 63.3 & 65.5 & 73.4\\
		w/o commitment repair & 65.4 & 66.2 & 74.1\\
		\bottomrule
	\end{tabular}
	\caption{Ablation study on success rate (\%).}
	\label{tab:ablation}
\end{table}

\begin{table}[ht]
	\centering
	\begin{tabular}{lcccc}
		\toprule
		& \multicolumn{2}{c}{Shopping} & \multicolumn{2}{c}{Admin} \\
		\cmidrule(lr){2-3}\cmidrule(lr){4-5}
		Variant & Succ. & Time & Succ. & Time \\
		\midrule
		Hybrid testing & \textbf{67.6} & \textbf{219s} & \textbf{68.8} & \textbf{529s} \\
		Slow-only testing & 66.9 & 281s & 68.8 & 648s \\
		\bottomrule
	\end{tabular}
	\caption{Hybrid commitment testing reduces average per-task LLM call time without sacrificing success.}
	\label{tab:efficiency}
\end{table}

\begin{figure}[ht]
	\centering
	\includegraphics[width=0.75\columnwidth]{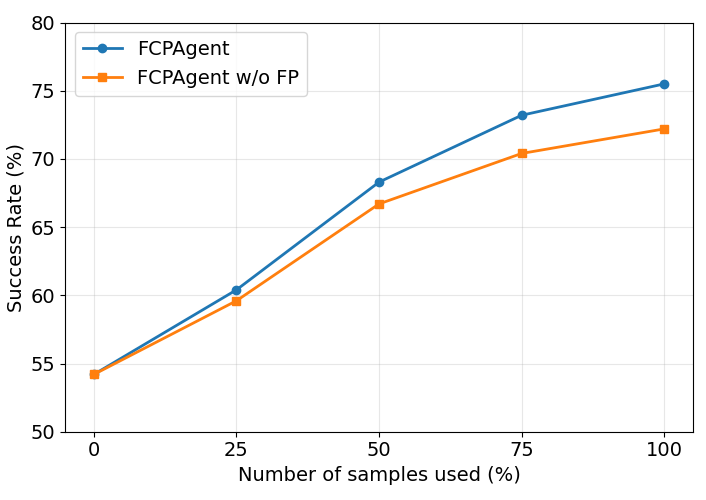}
	\caption{GitLab success rate as the amount of training experience increases. The top curve uses both the skill library and failure-repair library; the second curve removes the failure-repair library.}
	\label{fig:learning_curve}
\end{figure}

\begin{figure*}[ht]
	\centering
	\includegraphics[width=1.0\textwidth]{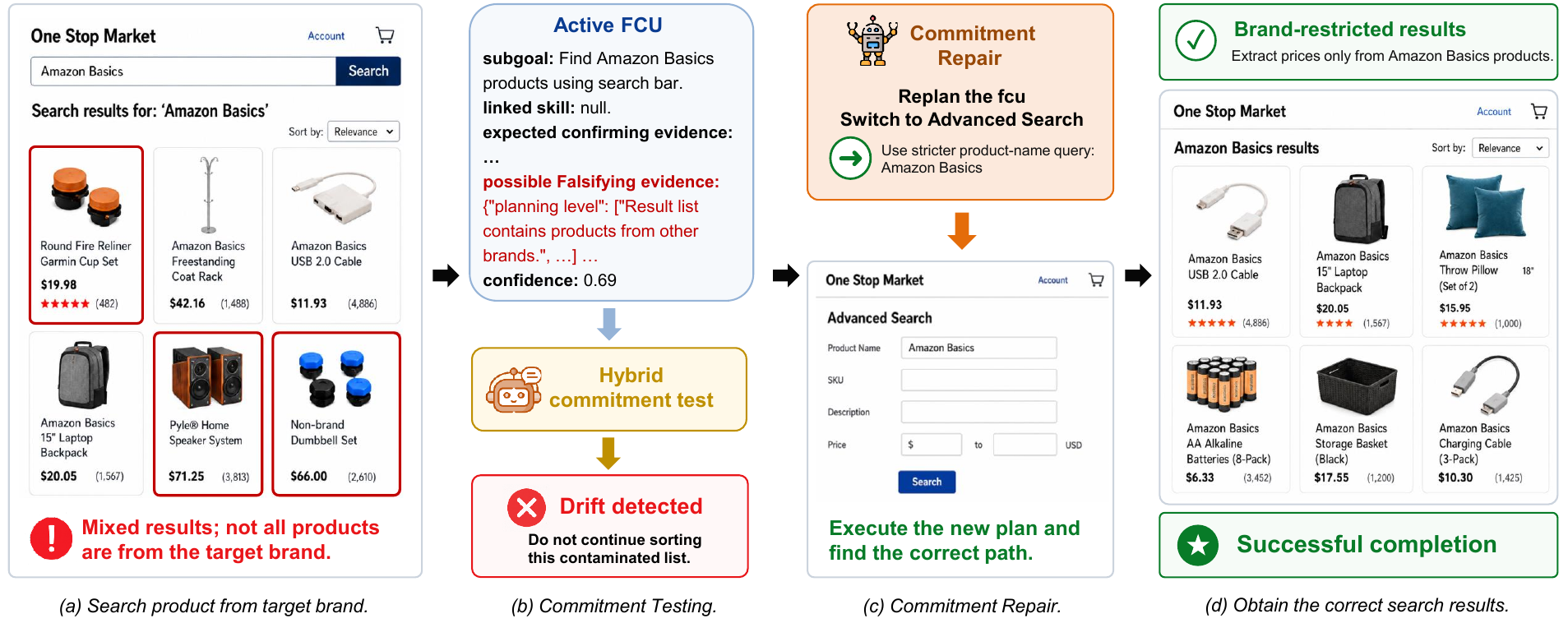}
	\caption{A WebArena-Shopping case study showing how \method{} detects plan-level drift, replans the active commitment, and obtains brand-restricted results.}
	\label{fig:case_study}
	\vspace{-2px}
\end{figure*}

\subsection{Ablation Study}

\subsubsection{Component ablation}
Table~\ref{tab:ablation} isolates the main components of \method{}. Across Shopping, Admin, and GitLab, the full model obtains a three-domain average success rate of 70.6\%. Removing falsifiable planning, where the executor receives only retrieved skills rather than generated \fcus{}, reduces the average to 64.4\%, the largest drop among the ablations. Removing hybrid commitment testing lowers the average to 67.4\%, confirming that \fcus{} are most useful when their evidence interface is checked online. Removing repair reduces the average to 68.6\%, indicating that explicit recovery contributes additional gains after evidence contradictions are detected. These results show that planning, testing, and repair play complementary roles: falsifiable planning supplies the tests, hybrid commitment testing applies them during execution, and scope-aware repair turns diagnosed contradictions into targeted revisions.

\subsubsection{Efficiency}
Table~\ref{tab:efficiency} evaluates the efficiency contribution of the lightweight evidence-matching layer. Replacing hybrid testing with slow-only verification yields similar success but increases total LLM call time. On average across Shopping and Admin, hybrid testing reduces per-task LLM call time from 464.5s to 374.0s, a 19.5\% reduction. This supports the hybrid design: LLM-based diagnostic verification is useful for ambiguous states, while lightweight tests handle routine transitions and reserve slow verification for moments that can change the commitment sequence.

\subsubsection{Experience scaling}
Figure~\ref{fig:learning_curve} studies how \method{} benefits from accumulated experience on GitLab. As the fraction of training trajectories used to build the experience libraries increases, the method variants improve steadily, showing that additional experience can be converted into better long-horizon execution. The full variant consistently outperforms the variant without the failure-repair library, indicating that failure memories provide complementary negative experience beyond reusable skills. This gap supports the role of failure-repair cases in helping the planner write sharper falsifiers and helping the repairer choose more effective scope-aware recoveries.

The Appendix provides finer-grained ablations of confirming and falsifying evidence and the action- and state-level testers, together with additional runtime efficiency and behavior analyses.

\subsection{Case Study}

Figure~\ref{fig:case_study} shows a WebArena-Shopping task that asks for the price range of Amazon Basics products. The agent first uses the ordinary search bar with the query \emph{Amazon Basics}, but the returned list is contaminated by products from other brands. Although the page appears relevant, continuing to sort this mixed list would produce a price range over non-target products.
Here, the broad search strategy in the active \fcu{} cannot satisfy the brand constraint because it returns products from other brands, constituting a plan-level drift. Hybrid commitment testing detects this evidence and prevents the agent from sorting the contaminated list. Scope-aware repair then replans the \fcu{}, switches to \emph{Advanced Search}, and issues a stricter product-name query. The resulting list contains only Amazon Basics products, allowing the agent to extract the correct price range. This case shows that falsifiable planning can identify not only a wrong execution path, but also an inadequate plan that must be revised.

\section{Conclusion}

We presented \method{}, a falsifiable commitment planning framework for long-horizon web agents. By representing plan steps as commitments with explicit confirming and falsifying evidence, \method{} changes execution from unconditional plan following into online plan-step testing. Experience-guided generation turns prior successes and failures into explicit confirming and falsifying evidence, hybrid commitment testing checks candidate actions and returned states against the active commitment, and scope-aware repair uses the violated evidence type to choose the smallest effective recovery. Experiments on WebArena show consistent gains over strong baselines, particularly on tasks with longer interaction horizons. These results suggest that stable web agents need not only better actions and richer memories, but plans that can recognize when they are wrong.

A limitation of the current framework is that it depends on the quality of generated evidence: overly broad falsifiers may trigger unnecessary repairs, whereas overly narrow ones may miss off-track execution. Future work should explore learned falsifier generation and benchmarks that measure how early agents detect recoverable deviations, not only final success.


\bibliography{aaai2027}


\clearpage
\appendix

\onecolumn

\section{Discussion}

\subsection{Comparison with Related Work}
\label{app:related-comparison}

\begin{table*}[ht]
    \centering
    \small
    \caption{
        Comparison of representative agent frameworks for reliable
        long-horizon execution.
    }
    \label{tab:related_work_comparison_appendix}
    \renewcommand{\arraystretch}{1.10}
    \begin{tabular}{lcccc}
        \toprule
        \textbf{Method}
        &\makecell{\textbf{Explicit Task-}\\\textbf{Progress Criteria}}
        &\makecell{\textbf{Explicit Drift}\\\textbf{Detection}}
        &\makecell{\textbf{Multi-Scope Failure}\\\textbf{Attribution}}
        &\makecell{\textbf{Attribution-Guided}\\\textbf{Recovery}}
        \\
        \midrule
        Reflexion
        & \xmark & \xmark & \xmark & \xmark \\
		Anticipatory Reflection
        & \xmark & \cmark & \xmark & \xmark \\
        AgentOccam
        & \xmark & \xmark & \xmark & \xmark \\
		AWM 
		& \xmark & \xmark & \xmark & \xmark \\
        WMA
        & \xmark & \xmark & \xmark & \xmark \\
        WebDART
        & \xmark & \cmark & \xmark & \xmark \\
        WALT
        & \xmark & \xmark & \xmark & \xmark \\
        SkillTracer
        & \xmark & \cmark & \xmark & \cmark \\
        ContractSkill
        & \xmark & \cmark & \xmark & \cmark \\
        ColorBrowserAgent
        & \xmark & \cmark & \xmark & \xmark \\
        \textbf{FCPAgent (Ours)}
        & \cmark & \cmark & \cmark & \cmark \\
        \bottomrule
    \end{tabular}
\end{table*}

Table~\ref{tab:related_work_comparison_appendix} evaluates representative agent frameworks along four requirements that follow the reliability process of specifying valid progress, detecting deviation, attributing its source, and selecting a corresponding recovery. \textbf{Explicit task-progress criteria (ETPC)} require task-specific prospective conditions that indicate whether the active plan step is progressing normally or has become invalid; generic skill preconditions or completion checks alone do not satisfy this criterion. \textbf{Explicit drift detection (EDD)} requires a dedicated mechanism for identifying deviation during the current trajectory rather than post-hoc reflection after task failure. \textbf{Multi-scope failure attribution (MSFA)} requires distinguishing whether a problem arises from immediate execution, the reused skill, or the higher-level plan. \textbf{Attribution-guided recovery (AGR)} requires the diagnosed source to explicitly determine the selected in-trajectory repair.

Reflexion~\cite{shinn2023reflexion} provides post-hoc self-evaluation, while AgentOccam~\cite{yang2025agentoccam}, AWM~\cite{wang2025awm}, WMA~\cite{chae2025wma}, and WALT~\cite{prabhu2026walt} improve action abstraction, outcome prediction, or procedure reuse. These capabilities can support execution, but they do not provide dedicated online drift detection under the definition above.

Anticipatory Reflection (AR)~\cite{wang2024devils} generates backup actions before execution and checks subtask alignment afterward, thus providing explicit online drift detection. However, its anticipation concerns alternative actions rather than task-specific validity criteria, and it does not attribute failures across multiple scopes to guide recovery.

WebDART~\cite{yang2025webdart} updates its task decomposition when execution no longer supports the current plan, and ColorBrowserAgent~\cite{zhou2026colorbrowseragent} combines progress assessment with environment adaptation. Both therefore provide explicit drift detection, but neither specifies task-specific prospective validity criteria for each active step or attributes failures across execution, skill, and planning scopes.

SkillTracer~\cite{li2026skilltracer} and ContractSkill~\cite{lu2026contractskill} detect violations within structured skill representations and use the localized violation to guide targeted repair. Their attribution, however, remains within the skill representation rather than distinguishing among immediate execution, the reused skill, and the higher-level plan. They therefore satisfy EDD and AGR, but not MSFA or task-specific ETPC.

\method{} covers all four requirements by attaching task-specific confirming and falsifying evidence to each active \fcu{}, detecting violations during execution, attributing them across execution, skill, and planning scopes, and selecting recovery from the attributed source. Its distinction lies not in drift detection or localized repair alone, but in unifying task-specific validity criteria, multi-scope attribution, and attribution-guided recovery. We do not report numerical comparisons with SkillTracer or ContractSkill because neither provides a runnable implementation compatible with our WebArena evaluation setting.

\section{Implementation Details}

\subsection{Runtime Algorithm}
\label{app:algorithm}

\begin{algorithm}[ht]
	\caption{\method{} Runtime Loop}
	\label{alg:fcpagent}
	\begin{algorithmic}[1]
		\REQUIRE Task $q$, initial observation $o_0$, skill library $\mathcal{S}$, failure-repair library $\mathcal{F}$, step budget $B$
		\ENSURE Executed trajectory $\tau$ and final observation
		\STATE Retrieve skills $S_q^K \leftarrow R_{\mathcal{S}}^K(q)$ and failure memories $F_q^P \leftarrow R_{\mathcal{F}}^P(q)$
		\STATE Generate falsifiable commitments $C=(c_1,\ldots,c_m) \sim \Pi(q,o_0,S_q^K,F_q^P)$
		\STATE Initialize $i\leftarrow 1$, $t\leftarrow 0$, and $\tau_{\le 0}\leftarrow(o_0)$
		\WHILE{$i\le m$ and $t<B$}
		\STATE Set active commitment $c_i\leftarrow C_i$
		\STATE Executor proposes reasoning and action $(r_t,a_t)$ from $(q,c_i,\tau_{\le t},o_t)$
		\STATE Run pre-action fast commitment test to obtain route $z_t^{\mathrm{pre}}$
		\IF{$z_t^{\mathrm{pre}}\ne\mathrm{continue}$}
		\STATE $(d_t,u_t)\leftarrow \Omega(q,c_i,\tau_{\le t},o_t,a_t,z_t^{\mathrm{pre}})$
		\IF{$d_t=\mathrm{repair}$}
		\STATE $(\tilde c_i,\tilde C_{>i})\leftarrow \Gamma(\tau_{\le t},c_i,u_t,S_q^k,R_{\mathcal{F}}^p(q,u_t))$
		\STATE Replace $(c_i,C_{>i})$ with $(\tilde c_i,\tilde C_{>i})$ and continue
		\ENDIF
		\ENDIF
		\STATE Execute $a_t$, observe $o_{t+1}$, and append $(a_t,o_{t+1})$ to $\tau$
		\STATE Run post-state fast commitment test to obtain route $z_t^{\mathrm{post}}$
		\IF{$z_t^{\mathrm{post}}\ne\mathrm{continue}$}
		\STATE $(d_t,u_t)\leftarrow \Omega(q,c_i,\tau_{\le t+1},o_{t+1},\varnothing,z_t^{\mathrm{post}})$
		\IF{$d_t=\mathrm{repair}$}
		\STATE $(\tilde c_i,\tilde C_{>i})\leftarrow \Gamma(\tau_{\le t+1},c_i,u_t,S_q^k,R_{\mathcal{F}}^p(q,u_t))$
		\STATE Replace $(c_i,C_{>i})$ with $(\tilde c_i,\tilde C_{>i})$
		\ELSIF{$d_t=\mathrm{advance}$}
		\STATE Summarize the completed stage and set $i\leftarrow i+1$
		\ENDIF
		\ENDIF
		\STATE $t\leftarrow t+1$
		\ENDWHILE
		\STATE In offline experience building, distill successful trajectories into $\mathcal{S}$ and failed or repaired trajectories into $\mathcal{F}$
		\RETURN $\tau$
	\end{algorithmic}
\end{algorithm}

We present the runtime loop of \method{} in Algorithm~\ref{alg:fcpagent}. The agent first retrieves relevant skills and failure-repair cases, then generates a sequence of falsifiable commitments. It executes each commitment in order, using hybrid commitment testing to check both proposed actions and returned observations. If a violation is detected, the repairer revises the current commitment and any subsequent commitments as needed. The loop continues until all commitments are completed or the step budget is reached. After task execution, successful trajectories are distilled into the skill library, while failed or repaired trajectories are stored in the failure-repair library for future use.

\subsection{Example of a Falsifiable Commitment Sequence}

\begin{figure}[ht]
	\centering
	\includegraphics[width=.6\columnwidth]{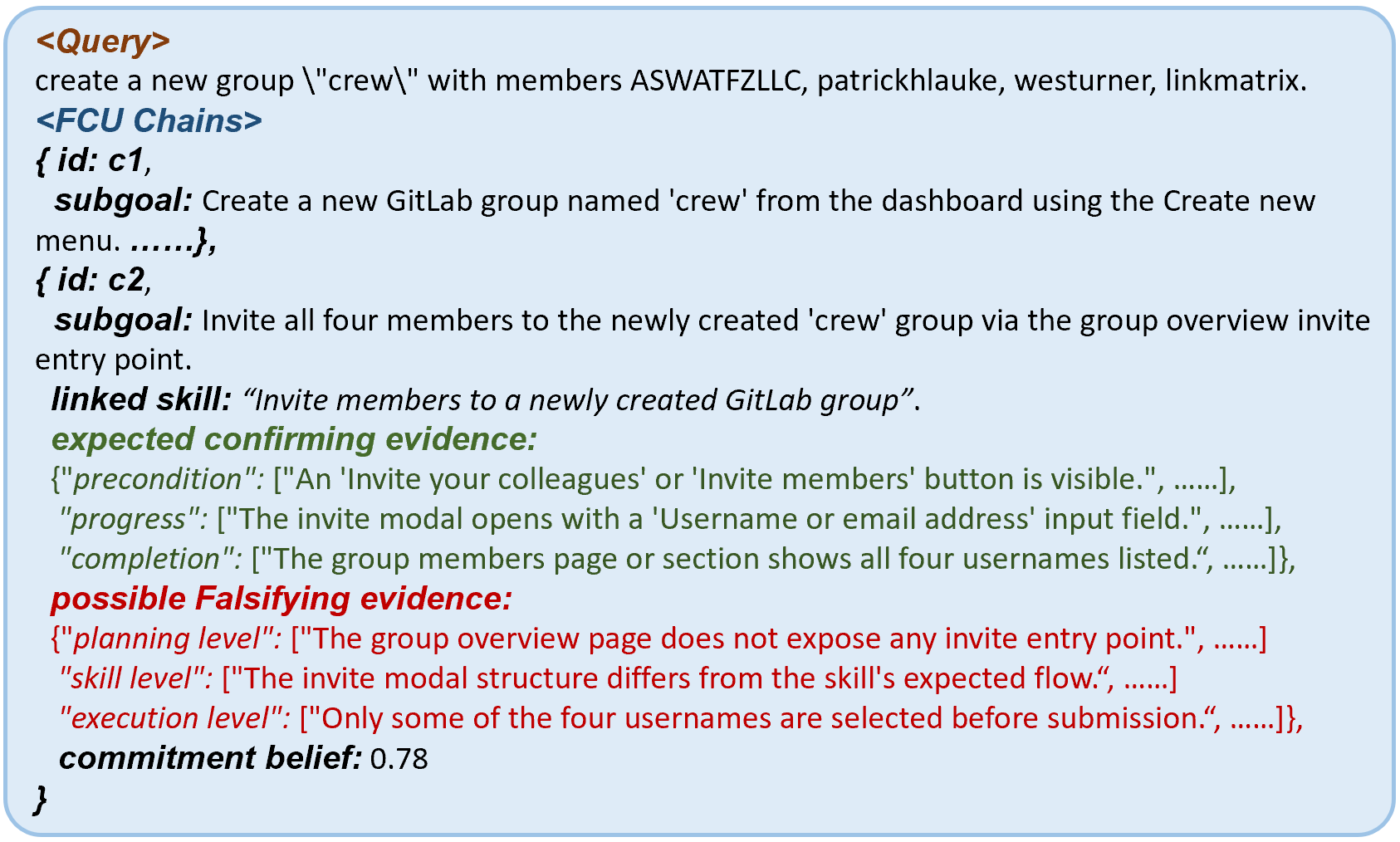}
	\vspace{-8px}
	\caption{Example of a falsifiable commitment sequence.}
	\label{fig:fcu-case}
\end{figure}

Figure~\ref{fig:fcu-case} shows an \fcu{} sequence for creating a GitLab group and inviting four members. The task is decomposed into two subgoals: creating the group and inviting its members. Each \fcu{} specifies a subgoal, an optional linked skill, stage-wise confirming evidence, hierarchical falsifying evidence, and a commitment confidence, providing explicit signals for runtime testing and repair. The confidence $\kappa_i$ is generated by the planner LLM to indicate its confidence in the accuracy of the current \fcu{}. It calibrates the fast-tester thresholds and is provided to the slow tester and repairer as a cue for deciding whether the \fcu{} itself should be revised.

\subsection{Implementation Details of the Skill Library}
\label{app:skill_library}

\begin{figure}[ht]
	\centering
	\includegraphics[width=.45\textwidth]{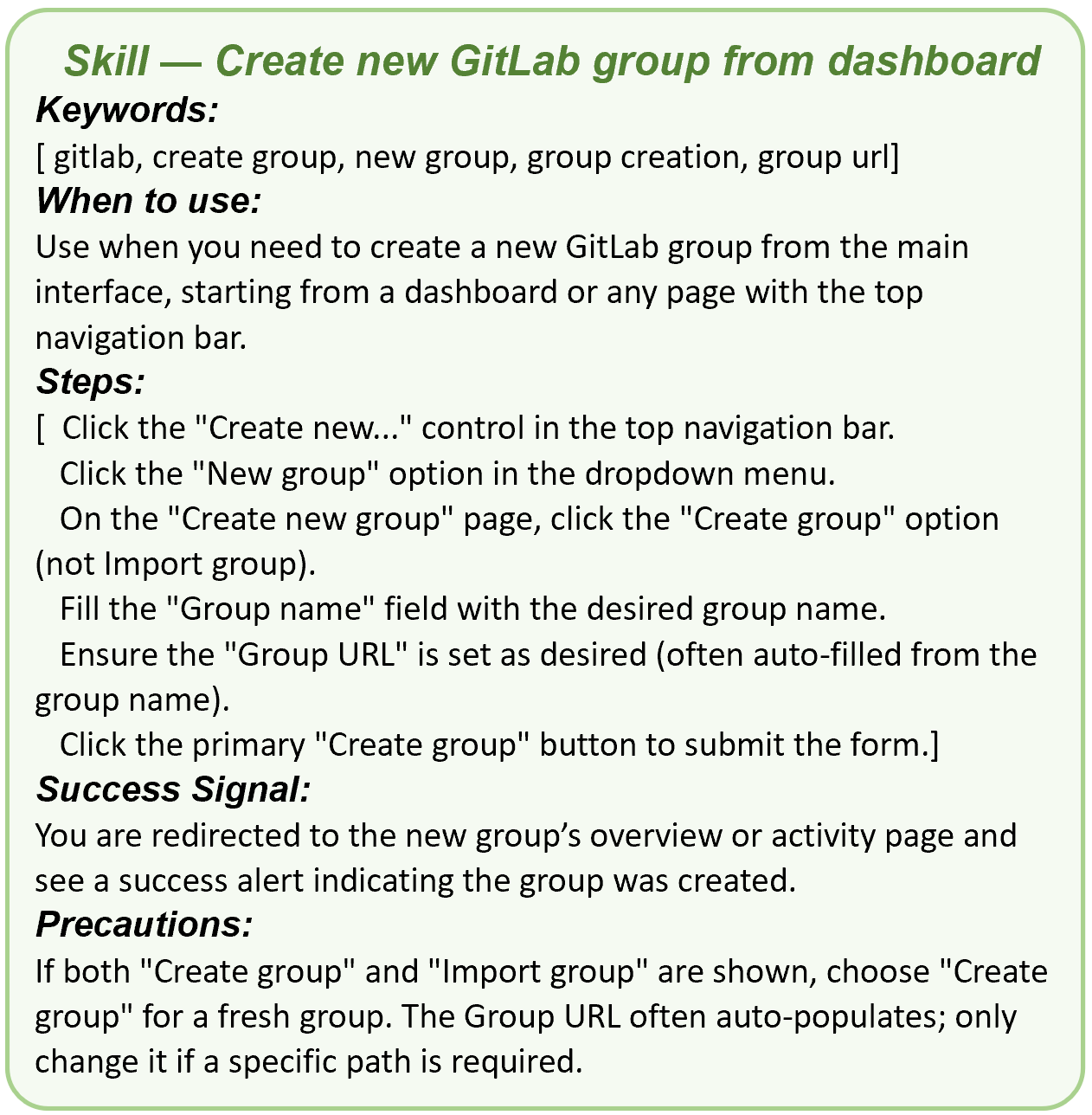}
	\caption{Example record from the skill library.}
	\label{fig:skill_library}
\end{figure}

FCPAgent maintains a unified skill library. Formally, each distilled skill is stored as
\begin{equation*}
s=(n,k,a,p,v,h),
\end{equation*}
where $n$ is the skill name, $k$ are retrieval keywords, $a$ describes applicability conditions, $p$ gives generalized execution steps, $v$ specifies observable success conditions, and $h$ records precautions. Records can be organized or filtered by environment metadata for efficient retrieval. As illustrated in Figure~\ref{fig:skill_library}, this library supports initial commitment generation, task execution, and scope-aware repair.

\paragraph{Hybrid skill retrieval.}
At the beginning of each task, FCPAgent retrieves skills using the user goal as the query. The retriever combines lexical and semantic similarity. For lexical matching, each skill is represented by its name, keywords, and applicability field, and a BM25 index is built over the tokenized skill texts. For semantic matching, each skill is represented by a richer text containing its name, keywords, applicability condition, and success signal. FCPAgent encodes both the query and skill texts with a local BGE-M3 encoder and compares embeddings by cosine similarity. 
The final retrieval score is a weighted combination of min-max-normalized BM25 and embedding scores:
\begin{equation}
	\sigma_{\mathcal{S}}(q,z)=0.4\,s_{\mathrm{BM25}}(q,z)+0.6\,s_{\mathrm{emb}}(q,z),
\end{equation}
where $q$ is the task goal and $z$ is a candidate skill. The top-$K$ skills under this score instantiate $R_{\mathcal{S}}^K(q)$ in the main text. 

\paragraph{Dependency expansion.}
The skill context is not limited to the independently top-ranked skills. After top-$K$ retrieval, FCPAgent expands the set with related skills induced from the same source task, adding nearby prerequisite or follow-up skills when available. These expansion items are tagged separately from directly retrieved skills. Thus, the final context may contain more than $K$ skills, combining top-ranked matches with related skills that preserve useful local workflow structure.

\paragraph{Use in planning and repair.}
The retrieved and expanded skill records are rendered into textual form and kept fixed for the task episode. The planner receives this candidate skill set when constructing the FCU chain and is instructed to link each FCU to at most one provided skill, using the skill's steps and success signal to write more concrete confirming evidence and disconfirming signals. The same skill set is reused during repair: when a commitment is falsified, the repair controller receives the current FCU, violation diagnosis, recent action history, and skill context, enabling scope-aware repair actions such as switching skills, rewriting the current FCU, or replanning the remaining commitments. During execution, the low-level action generator receives only the active skill attached to the current FCU, keeping skill guidance localized. The implementation also supports partial or progressive disclosure of skill details, but the standard configuration directly provides the retrieved skill contents.

\paragraph{Skill induction.}
Skills are induced offline from successful WebArena trajectories. For each successful task, an LLM extracts at most five reusable skills from the task goal and execution trajectory. The prompt emphasizes generality rather than memorization by removing raw DOM ids, coordinates, and action argument ids while preserving operational steps. It encourages splitting a trajectory only when the split improves reuse, with emphasis on navigation skills, information-extraction skills, and reasoning rules.

\subsection{Implementation Details of the Failure-Repair Library}
\label{app:failure_repair_library}

\begin{figure}[ht]
	\centering
	\includegraphics[width=.45\textwidth]{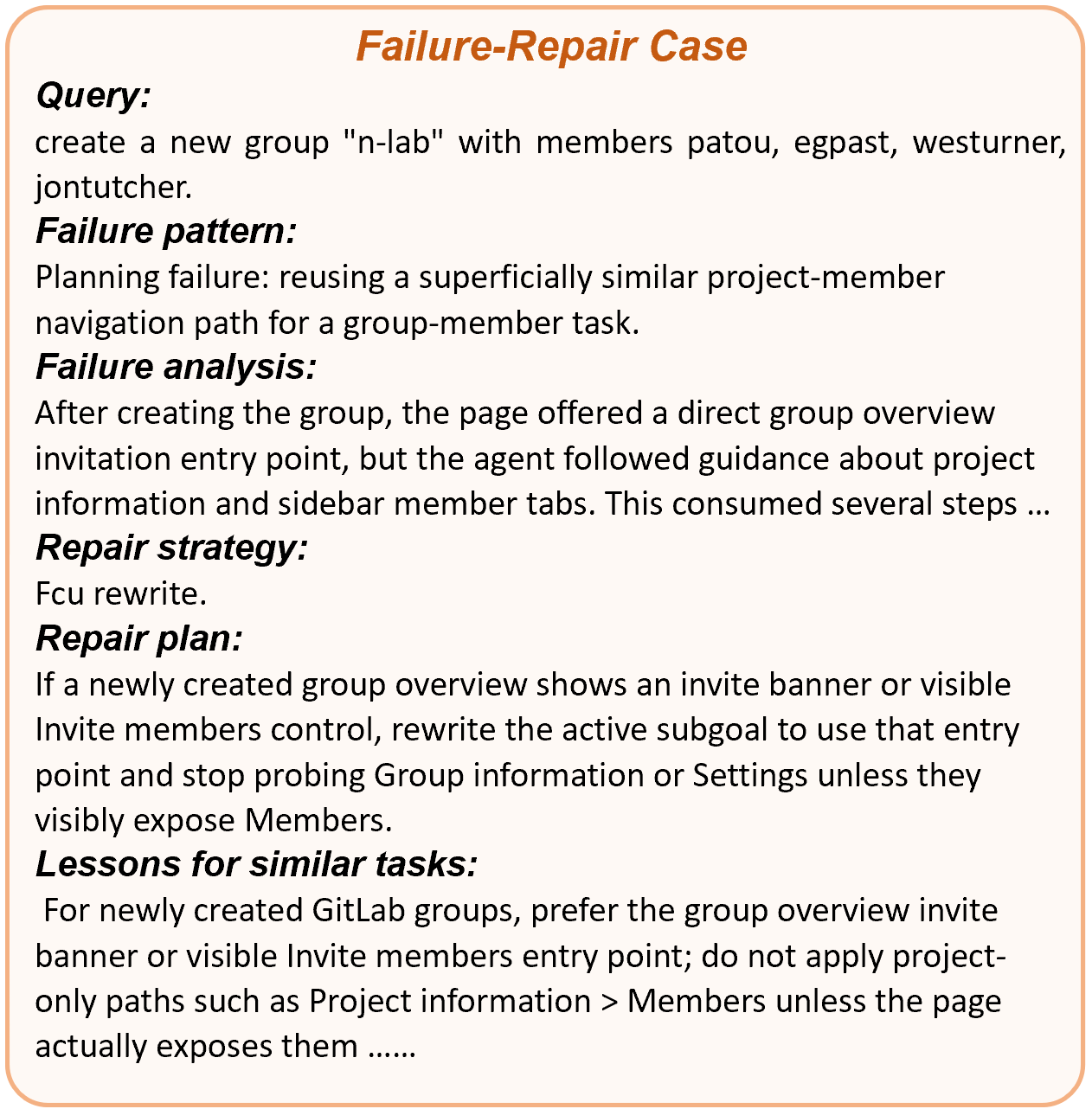}
	\caption{Example record from the failure-repair library.}
	\label{fig:failure_repair_library}
\end{figure}

FCPAgent also maintains a unified failure-repair library built from failed or repaired task executions. A failure-repair case is stored as
\begin{equation*}
f=(q,\rho,\gamma,\eta),
\end{equation*}
where $q$ is the original task query, $\rho$ contains the failure pattern and evidence-grounded analysis, $\gamma$ specifies the repair strategy and plan, and $\eta$ records lessons for similar tasks. As shown in Figure~\ref{fig:failure_repair_library}, each record provides compact repair-oriented experience. Distinct failure modes from one trajectory are stored as separate cases rather than merged.

\paragraph{Retrieval and use.}
During execution, each stored case is represented with two semantic views: one for the original task query and one for the failure pattern and analysis. At repair time, FCPAgent retrieves cases using both the current task query $q$ and the current violation diagnosis $u$, matching the repair input $R_{\mathcal{F}}^P(q,u_t)$ in the main text. The final score is
\begin{equation}
	\sigma_{\mathcal{F}}(q,u,r)=0.5\,s_{\mathrm{query}}(q,r)+0.5\,s_{\mathrm{diag}}(u,r),
\end{equation}
where $r$ is a stored failure-repair record. Embeddings are computed with the same local semantic encoder used by the skill retriever and cached for reuse. Retrieved cases are filtered by a similarity threshold and truncated to the configured top-$P$ results.

Before planning, FCPAgent may retrieve failure-repair cases using only the task query, providing high-level lessons that help the planner specify stronger falsifiers and avoid known failure modes. When a commitment is falsified during execution, the repair controller retrieves cases using both the task query and the current violation diagnosis. The retrieved records are injected as soft guidance containing the original query, similarity score, failure pattern, failure analysis, repair strategy, and repair plan. They do not override the current browser state or active FCU; instead, they bias scope-aware repair toward remedies that were useful in similar failures.

\paragraph{Failure-repair induction.}
The library is constructed offline. For each failed run, FCPAgent reads the task query and recorded planner-action trajectory, then asks an LLM to summarize the failure into concise, reusable repair experience. The failure pattern is written at an abstract level, while the failure analysis remains tied to concrete observed behavior. The repair strategy is normalized to one of the supported repair actions, such as local repair, skill switching, backtracking, FCU rewriting, or global replanning. The repair plan gives actionable guidance for future repair decisions without hard-coding a recovery action.

\subsection{Hybrid Commitment Testing Details}
\label{app:commitment-testing}

\method{} implements hybrid commitment testing with a lightweight fast tester followed by an LLM-based slow verifier. The fast tester does not make final repair decisions; it routes ordinary states to \texttt{continue} and escalates likely completion, likely drift, execution anomalies, or uncertain boundary cases to the slow verifier.

\paragraph{State-level fast tester.}
After each browser action, the state-level tester evaluates the active \fcu{} against the current observation. Its inputs include the active commitment, grouped confirming and falsifying evidence, the latest action status, open-tab URLs and titles, the accessibility tree, and the screenshot. Before semantic scoring, it applies rule checks for obvious anomalies, such as browser execution errors, unparsable actions, unchanged observations after page-changing actions, or web error pages. These cases are routed directly to slow verification as possible execution-level drift.

If no rule fires, the tester serializes the page context from tab metadata and the accessibility tree and chunks long observations. Let $T_t$ and $I_t$ denote the serialized text and screenshot, and let $E_i^{\pm}$ denote confirming evidence $\pos_i$ or falsifying evidence $\fals_i$. Their matching scores are
\begin{equation}
\alpha_t^{\pm}=\lambda\,\mathrm{NLI}(T_t,E_i^{\pm})+(1-\lambda)\,\mathrm{ITM}(I_t,E_i^{\pm}).
\label{eq:app-evidence-match}
\end{equation}
Textual evidence is matched with a DeBERTa NLI cross-encoder, taking the maximum entailment score over chunks, while SigLIP provides image--text similarity. Confirming evidence is scored over progress and completion signals, whereas falsifying evidence is scored separately over planning-, skill-, and execution-level groups. The highest-scoring falsifier group is retained as a diagnostic hint for repair.

Let $\alpha_{t,\mathrm{comp}}^+$ denote the score of the completion component of $\pos_i$. The fast tester routes the state as
\begin{equation}
z_t=
\begin{cases}
\mathrm{complete}, & \alpha_{t,\mathrm{comp}}^+ \ge \tau_+(\kappa_i)\ \wedge\ \alpha_{t,\mathrm{comp}}^+-\alpha_t^- \ge \delta_+,\\
\mathrm{risk}, & \alpha_t^- \ge \tau_-(\kappa_i)\ \wedge\ \alpha_t^- - \alpha_t^+ \ge \delta_-,\\
\mathrm{continue}, & \max(\alpha_t^+,\alpha_t^-) < \pi_{\mathrm{low}},\\
\mathrm{verify}, & \text{otherwise}.
\end{cases}
\label{eq:app-fast-routing}
\end{equation}
The \texttt{complete}, \texttt{risk}, and \texttt{verify} routes invoke the slow verifier for completion-like, drift-like, and ambiguous cases, respectively; they are routing cues rather than final decisions. Only \texttt{continue} skips the LLM call. In the WebArena runner, the base gates are $\tau_+^0=0.65$ and $\tau_-^0=0.55$, adjusted by confidence as $\tau_\pm(\kappa_i)=\tau_\pm^0+\eta(\kappa_i-0.5)$ with $\eta=0.1$. We set $\pi_{\mathrm{low}}=0.5$, use $\delta_+=0.18$ for completion and $\delta_-=0.25$ for drift, and apply an uncertainty margin of $0.08$ to near-threshold cases. Lower-confidence commitments are therefore escalated more readily. A periodic slow check after several consecutive \texttt{continue} decisions prevents weak signals from being ignored indefinitely.

\paragraph{Action-level fast tester.}
Before executing selected high-impact actions, \method{} optionally runs an action-level fast tester. It builds an action description from the proposed browser command, the executor reasoning, the active \fcu{}, and the resolved accessibility-tree target when the action refers to an element id. The tester applies the textual component of Equation~\ref{eq:app-evidence-match}, replacing $T_t$ with this action description and omitting the visual term. It then estimates whether the action aligns with the active subgoal and progress evidence and whether it overlaps with falsifying evidence that can be judged before execution. Observation-only falsifiers are filtered out because they require the post-action page state.

The resulting action-risk score emphasizes overlap with falsifying evidence more than lack of positive alignment. Low-risk actions are executed normally. High-risk actions are not blocked by the fast tester alone; instead, they are sent to the slow verifier together with the current page, active \fcu{}, proposed action, and fast-test warning. If the slow verifier rejects the action, repair is triggered before the browser state changes, and the rejected action is recorded to discourage immediate repetition.

Together, the state-level and action-level fast testers provide a low-cost evidence-matching layer around the LLM verifier. This preserves the plan-test-repair loop while reducing unnecessary LLM calls on routine browser transitions.

\subsection{Dataset Split and Evaluation Protocol}
\label{app:dataset_split}

\begin{table}[ht]
	\centering
	\begin{tabular}{lccccccc}
		\toprule
		Split & Shopping & Admin & GitLab & Reddit & Map & Cross & Total \\
		\midrule
		Train & 48  & 41  & 41  & 21 & 29 & 11 & 191 \\
		Test  & 139 & 141 & 139 & 85 & 80 & 37 & 621 \\
		\midrule
		Total & 187 & 182 & 180 & 106 & 109 & 48 & 812 \\
		\bottomrule
	\end{tabular}
	\caption{Dataset split used for offline experience construction and test evaluation.}
	\label{tab:dataset_split}
\end{table}

Table~\ref{tab:dataset_split} reports the dataset split used in our experiments. The benchmark contains six WebArena domains, including Shopping, Admin, GitLab, Reddit, Map, and Cross-domain tasks. We use 191 tasks for offline experience construction and 621 tasks for final evaluation.

The training split is used only to construct the offline experience support for FCPAgent. Specifically, we build two fixed experience libraries from the training tasks: a skill library distilled from successful trajectories and a failure-repair library distilled from failed or repaired trajectories. During test-time evaluation, both libraries are kept frozen, and no test trajectory is used to update, expand, or modify the experience libraries.

All methods are evaluated on the same test split. To reduce variance caused by stochastic decoding and interaction trajectories, each method is run three times on the test set, and we report the average success rate across the three runs.

\subsection{Baselines}

We compare against six representative web-agent baselines that cover reactive prompting, workflow memory, observation/action abstraction, dynamic replanning, tool induction, and long-horizon progress adaptation.
\begin{itemize}[leftmargin=*]
	\item \textbf{ReAct}~\cite{yao2023react} interleaves natural-language reasoning traces with environment actions. We use it as a general prompting baseline for step-by-step web interaction.
	\item \textbf{AWM}~\cite{wang2025awm} induces reusable workflows from past trajectories and retrieves relevant workflows to guide future tasks. It represents the memory-based reuse line of web-agent methods.
	\item \textbf{AgentOccam}~\cite{yang2025agentoccam} improves LLM-based web agents by simplifying and aligning the observation and action spaces with the model's capabilities. It is a strong zero-shot web-agent baseline without explicit workflow memory or repair.
	\item \textbf{WebDART}~\cite{yang2025webdart} dynamically decomposes complex web tasks into navigation, information extraction, and execution subtasks, and replans as new pages are observed.
	\item \textbf{WALT}~\cite{prabhu2026walt} learns callable tools that expose recurring website functions such as search, filtering, sorting, posting, or editing. These tools compress multi-step browser interactions into higher-level operations.
	\item \textbf{ColorBrowserAgent}~\cite{zhou2026colorbrowseragent} targets complex long-horizon web automation with progress summarization and environment adaptation. It is the strongest baseline in our comparison.
\end{itemize}

\subsection{Implementation Details}

We use Qwen3.5-397B-A17B as the backbone model \cite{qwen2026qwen35} via API call for all methods, with the temperature set to 1. We typically set the number of retrieved skills $K = 3$, the number of retrieved failure experiences $P = 2$, and the hyperparameter $\lambda = 0.8$.

Following ColorBrowserAgent~\cite{zhou2026colorbrowseragent}, our system is implemented on top of BrowserGym~\cite{chezelles2024browsergym}, with Playwright for programmatic browser control. Observations consist of multimodal inputs, combining Set-of-Marks~\cite{yang2023setofmark} and accessibility-tree representations from WebArena~\cite{zhou2024webarena}. We adopt an action space adapted from AgentOccam~\cite{yang2025agentoccam} to reduce execution errors. The maximum interaction length is set to 20 steps.

\section{Complementary Experiments}

\subsection{Efficiency Analysis}
\label{app:efficiency_analysis}

\begin{figure}[ht]
	\centering
	\begin{subfigure}[t]{.48\textwidth}
		\centering
		\includegraphics[width=\linewidth]{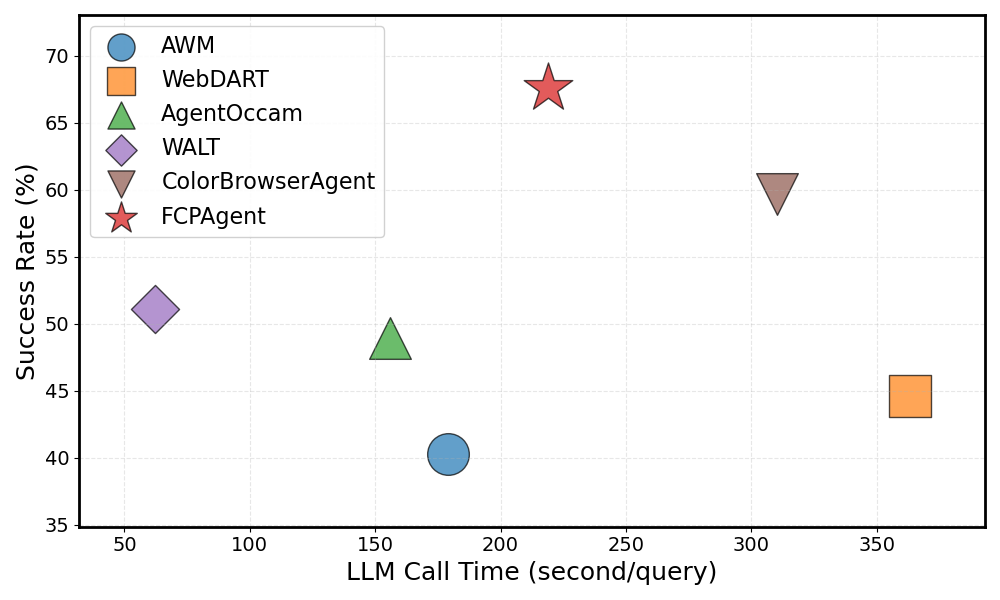}
		\caption{LLM call time.}
		\label{fig:time_llm}
	\end{subfigure}
	\hfill
	\begin{subfigure}[t]{.48\textwidth}
		\centering
		\includegraphics[width=\linewidth]{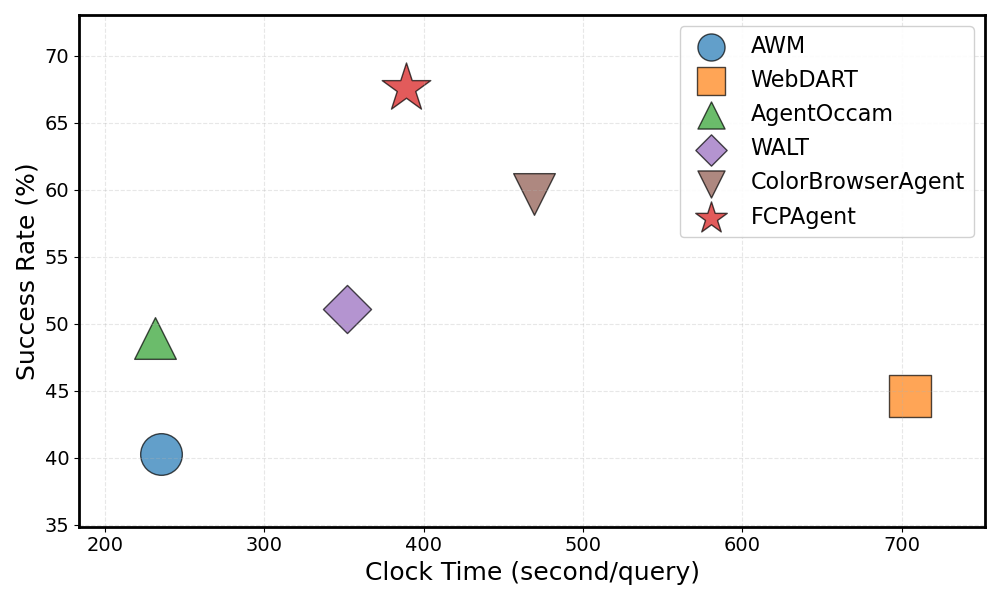}
		\caption{End-to-end clock time.}
		\label{fig:time_clock}
	\end{subfigure}
	\caption{Efficiency and success rate on Shopping tasks under two runtime measures. The y-axis reports task success rate, while the x-axis reports the average time per query.}
	\label{fig:time}
\end{figure}

Figure~\ref{fig:time} compares the efficiency--performance trade-off on Shopping tasks using two complementary runtime measures. Figure~\ref{fig:time_llm} reports LLM call time, which isolates the dominant model-inference cost. Methods that rely heavily on repeated planning and reflection incur substantially higher LLM latency: ColorBrowserAgent requires more than 300 seconds per query, while WebDART is even slower. In contrast, \method{} achieves the highest success rate while keeping LLM call time at a moderate level. ColorBrowserAgent~\cite{zhou2026colorbrowseragent} is a particularly relevant comparison because it is a closely related reflection-based baseline: after every executed action, it reflects on task progress and whether the trajectory has drifted. In our runtime profiling, this reflection stage consumes more time than its action executor, explaining much of its latency. This observation is consistent with the module-level analysis of \method{} in Figure~\ref{fig:module_runtime}, where slow LLM verification also costs more than execution, and motivates avoiding unconditional reflection after every action.

Figure~\ref{fig:time_clock} reports end-to-end clock time, which additionally includes browser loading, webpage-environment response, tool execution, and other non-LLM overheads. This broader measure changes the apparent efficiency of WALT most noticeably. Although its induced tools reduce LLM reasoning time by compressing recurring routines, executing these tools requires frequent interactions with the webpage, causing WALT to move substantially to the right in the clock-time comparison. By contrast, the relative position of \method{} changes little across the two views. Its fast tester is lightweight and inexpensive, as further quantified by the module-level runtime analysis below, while slow LLM verification is reserved for completion-like, risky, or ambiguous cases. Consequently, hybrid commitment testing improves reliability without introducing disproportionate end-to-end latency, and \method{} retains a favorable efficiency--success trade-off under both runtime measures.

\subsection{Fine-grained evidence and testing ablations}

\begin{table}[ht]
	\centering
	\begin{tabular}{lcc}
		\toprule
		Variant & Shopping & Admin  \\
		\midrule
		Full  & \textbf{67.6} & \textbf{68.8}\\
		w/o confirming evidence & 64.7 & 66.0\\
		w/o falsifying evidence & 62.5 & 63.8\\
		w/o state-level test & 64.0 & 66.0 \\
		w/o action-level test & 66.9 & 68.1\\
		w/o local repair & 64.7 & 65.2\\
		w/o replan & 65.4 & 67.4\\
		\bottomrule
	\end{tabular}
	\caption{Fine-grained ablation study on success rate (\%).}
	\label{tab:fine_grained_ablation}
\end{table}

Table~\ref{tab:fine_grained_ablation} further decomposes the contributions of the evidence representation, the two runtime testing interfaces, and the available repair scopes. Removing confirming evidence reduces the average success rate from 68.2\% to 65.4\%, showing that prospective precondition, progress, and completion signals help the agent recognize whether the active commitment remains supported. Removing falsifying evidence causes the largest degradation, reducing the average to 63.2\%. This result supports the central motivation of FCPAgent: explicitly specifying observations that would invalidate the current action, skill, or plan is important for detecting locally plausible but task-irrelevant execution.

The testing ablations further show that pre-action and post-state checks play complementary but asymmetric roles. Removing post-state testing reduces the average success rate to 65.0\%, because many forms of trajectory drift become observable only after the browser returns a new state. In contrast, removing pre-action testing produces a smaller reduction to 67.5\%. This is consistent with the runtime diagnostics in Table~\ref{tab:runtime_behavior}, where state-level testing detects suspicious situations more frequently, while action-level testing acts as a lighter preventive guard before high-impact actions modify the environment. Overall, the results indicate that falsifying evidence and post-state testing provide the primary robustness gains, while confirming evidence and pre-action testing offer complementary support for progress assessment and preventive correction.

The repair-scope ablations show that recovery should not be restricted to a fixed scope. Disabling local repair reduces average success from 68.2\% to 65.0\%, while disabling replanning reduces it to 66.4\%. The larger drop without local repair reflects the value of preserving valid commitments when a deviation can be corrected locally, whereas the remaining degradation without replanning shows that some falsified commitments require revising the higher-level plan. Together, these results support scope-aware repair, which uses failure attribution to select the smallest adequate correction instead of defaulting to either local retry or global replanning.

\subsection{Runtime Behavior Analysis}
\label{sec:runtime_behavior}

\begin{table}[ht]
\centering
\begin{tabular}{lccccccccc}
\toprule
& \multicolumn{3}{c}{Plan} & \multicolumn{4}{c}{Test} & \multicolumn{2}{c}{Repair} \\
\cmidrule(lr){2-4}\cmidrule(lr){5-8}\cmidrule(lr){9-10}
Task & \shortstack{Tasks w/\\skill} & \shortstack{FCUs\\linked} & \shortstack{FCUs\\/task} & \shortstack{State\\slow call} & \shortstack{State\\drift} & \shortstack{Action\\slow call} & \shortstack{Action\\drift} & \shortstack{Repair\\/task} & \shortstack{Repair\\succ.} \\
\midrule
Shopping & 94\% & 78\% & 2.6 & 72\% & 10\% & 14\% & 6\% & 0.4 & 73\% \\ 
Admin & 90\% & 88\% & 2.9 & 77\% & 15\% & 13\% & 7\% & 1.0 & 45\% \\
GitLab & 93\% & 72\% & 2.8 & 77\% & 17\% & 18\% & 8\% & 0.9 & 40\% \\
\bottomrule
\end{tabular}
\caption{Runtime behavior of FCPAgent on three representative domains. Tasks w/ skill is the fraction of tasks covered by at least one retrieved skill; FCUs linked is the fraction of generated FCUs linked to skills. State/action slow call reports how often the corresponding fast tester routes to slow verification, while state/action drift reports the fraction of slow-verifier calls diagnosed as drift. Repair/task denotes the average number of drift-triggered repair events per task.}
\label{tab:runtime_behavior}
\end{table}

To better understand how FCPAgent behaves during execution, we additionally instrument the plan-test-repair loop on three representative domains. Table~\ref{tab:runtime_behavior} reports planning coverage, commitment-testing behavior, and repair outcomes. These diagnostics complement final task success by showing whether the proposed components are actually used during long-horizon execution.

The planning statistics show that the offline skill library provides broad procedural support: 90--94\% of tasks are covered by at least one relevant retrieved skill, and 72--88\% of generated FCUs are linked to a skill. Each task is decomposed into roughly three FCUs on average, suggesting that FCPAgent usually plans at a compact subgoal level rather than issuing a long flat action script. At the same time, skill coverage alone is insufficient for robust execution, since a retrieved skill may become invalid under a different page state, layout, or task-specific constraint.

The testing statistics reveal an asymmetry between state-level and action-level testing. State-level fast testing routes 72--77\% of cases to slow verification, and 10--17\% of these slow checks are diagnosed as drift. In contrast, action-level fast testing routes only 13--18\% of cases to slow verification, with 6--8\% diagnosed as drift. This pattern is expected for web tasks: risks are often easier to perceive from the resulting page state, while individual agent actions are frequently clicks or navigational moves whose risk is less explicit before execution. Thus, state-level testing provides higher recall for suspicious situations, whereas action-level testing acts as a lighter pre-action guard.

The repair statistics show that drift-triggered repair is used sparsely: 0.4 times per Shopping task and about once per Admin or GitLab task. Once repair is invoked, the eventual success rate is 73\% on Shopping and 40\% on Admin and GitLab. The lower repair-conditioned success in Admin and GitLab is expected because these domains often involve more stateful operations, permission-sensitive pages, and irreversible intermediate choices. Nevertheless, the results show that scope-aware repair can recover a non-trivial fraction of trajectories after drift has already been detected.

\subsection{Drift Detection Accuracy}
\label{app:drift_detection}

\begin{table}[ht]
	\centering
	\begin{tabular}{lccc}
		\toprule
		Manual annotation & Detected & Not detected & Total \\
		\midrule
		Drift present & 10 (TP) & 3 (FN) & 13 \\
		No drift & 2 (FP) & 15 (TN) & 17 \\
		\midrule
		Total & 12 & 18 & 30 \\
		\bottomrule
	\end{tabular}
	\caption{Task-level drift detection on 30 randomly sampled trajectories.}
	\label{tab:drift_detection}
\end{table}

To directly evaluate whether \method{} detects trajectory drift, we randomly sample 30 evaluated task trajectories, of which 20 are successful (66.7\%), and manually annotate whether each trajectory contains drift. At the task level, a drift-containing trajectory is counted as detected only when \method{} raises at least one diagnosis that manual inspection confirms corresponds to an actual drift event. Table~\ref{tab:drift_detection} compares these diagnoses with the manual annotations.

The detector achieves 83.3\% precision, 76.9\% recall, and an 80.0\% F1 score, with 83.3\% overall accuracy. The high precision indicates that most raised drift diagnoses correspond to genuine trajectory deviations, while the recall shows that the tester identifies most, but not all, of the drift cases found by manual inspection. These results provide direct evidence that falsifiable commitment testing detects meaningful execution drift, while also showing that subtle deviations remain an important target for improvement.

\subsection{Module-Level Runtime Analysis}
\label{app:module_runtime}

\begin{figure}[ht]
	\centering
	\includegraphics[width=.45\textwidth]{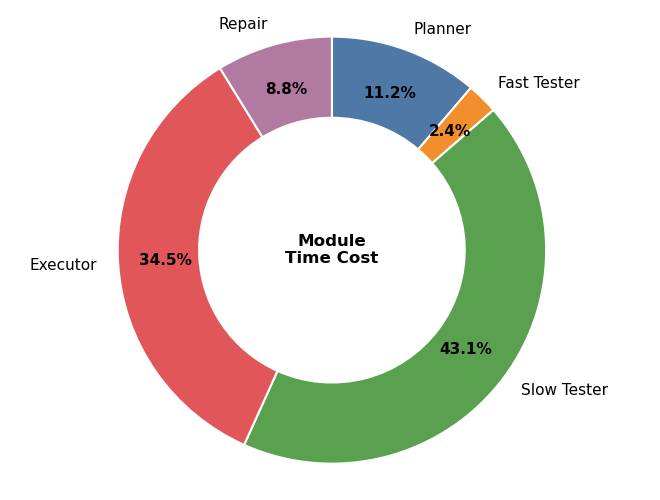}
	\caption{Distribution of test-time cost across the modules of \method{}.}
	\label{fig:module_runtime}
\end{figure}

Figure~\ref{fig:module_runtime} breaks down the total test-time cost across the major modules of \method{}. The fast tester accounts for only 2.4\% of the overall runtime, confirming that lightweight evidence matching can be applied frequently with little overhead. In contrast, the slow tester accounts for 43.1\%, making it the most expensive module and even exceeding the executor's 34.5\% share. Planning and repair contribute the remaining 11.2\% and 8.8\%, respectively. This distribution highlights the importance of hybrid commitment testing: invoking the slow LLM verifier after every browser transition would impose substantial unnecessary cost, whereas the fast tester can filter routine states and reserve slow diagnosis for completion-like, risky, or ambiguous cases.

\subsection{Repair Strategy Distribution}
\label{app:repair_distribution}

\begin{figure}[ht]
	\centering
	\includegraphics[width=.45\textwidth]{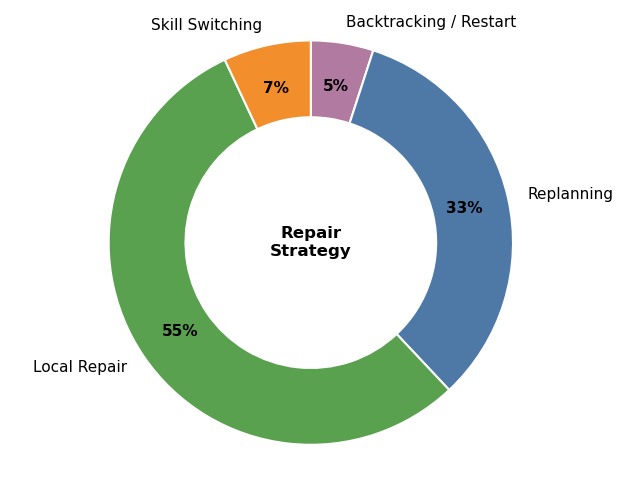}
	\caption{Distribution of repair strategies selected by the scope-aware repairer.}
	\label{fig:repair_distribution}
\end{figure}

Figure~\ref{fig:repair_distribution} shows how the repairer distributes its decisions across repair scopes. Local repair is selected most frequently (55\%), consistent with the design principle of revising the smallest contradicted component. Replanning nevertheless accounts for a substantial 33\%, indicating that many detected violations cannot be resolved by correcting only the next action and instead require revising the active or remaining commitments. Skill switching (7\%) handles cases in which the current subgoal remains valid but the retrieved procedure no longer fits the browser state, while backtracking or restarting (5\%) provides recovery from severe deviations or unsafe intermediate states. Although less frequent, these strategies show that no single repair scope is sufficient: all four address distinct failure conditions and jointly support heterogeneous long-horizon recovery.

\end{document}